# The Chess Transformer:
# Mastering Play using Generative Language Models

## David Noever[1], Matthew Ciolino[2], Josh Kalin[3]

PeopleTec,[1,2,3] Auburn University[3]
david.noever@peopletec.com,[1] matt.ciolino@peopletec.com,[2] jzk0098@auburn.edu[3]

## Abstract

This work demonstrates that natural language transformers can support more generic strategic modeling, particularly for text-archived games. In addition to learning natural language skills, the abstract transformer architecture can generate meaningful moves on a chessboard. With further fine-tuning, the transformer learns complex gameplay by training on 2.8 million chess games in Portable Game Notation. After 30,000 training steps, OpenAI's Generative Pre-trained Transformer (GPT-2) optimizes weights for 774 million parameters. This fine-tuned Chess Transformer generates plausible strategies and displays game formations identifiable as classic openings, such as English or the Slav Exchange. Finally, in live play, the novel model demonstrates a human-to-transformer interface that correctly filters illegal moves and provides a novel method to challenge the transformer's chess strategies. We anticipate future work will build on this transformer's promise, particularly in other strategy games where features can capture the underlying complex rule syntax from simple but expressive player annotations.

## Introduction

This research learns the rules of chess without direct expert intervention or heuristic guidance. Extending previous work on learning Go games with language transformers (Ciolino et al. 2020), the work benefits from large archives of chess notation in text and a game replay visualization tool. We combine the millions of chess games in text formats with the remarkable feature learning parts of the large GPT-2 model (Radford et al. 2018). Unlike a traditional sequence generator, the transformers support built-in parallelism and a directed attention mechanism to overweight key features. The original contributions of this research include generating plausible chess moves following the fine-tuning of the large GPT-2 transformer with its 774 million model parameters. A second innovation features a novel game interface where human players can challenge the transformer in live play. To take advantage of graphical processing units (GPU-acceleration) we host the shared games on Google's Colaboratory platform (Colab)[1].

In contrast to our previous exploration of transformers to play Go, chess has received considerable attention in language modeling. A Lisp chess engine (Penson)[2] applied a frequency map and Portable Game Notation (PGN) Mentor Database. The frequency map establishes conditional move probabilities for each forward move. To produce move probabilities from a board state vector (Ahle 2018), another text-trained chess engine (fastchess)[3] applied a popular text classification library as a one-layer plus soft-max model. When first open-sourced in 2016, their fastText classifier (Bojanowski et al. 2017) from Facebook AI Research was state-of-the-art. Facebook employed sentence structure features with bags of both words and n-grams, two strategies now overtaken in the literature by the rapid growth of transformers such as Google's BERT (Devlin et al. 2018) and OpenAI's GPT. These newer language models supplement the long tradition of using game trees to formalize decision-making and chess strategies (Nornai 1997). One can postulate that the decision tree model is deep (enumerating 60+ moves ahead) but narrow compared to the language-based alternatives presented by our chess transformers. This approach further contrasts with the Monte Carlo Tree Search, (MCTS) employed so effectively with Alpha-Go and reinforcement learning (Silver et al. 2018).

The application of models outside their initial language-related training sets has attracted interest in other cross-domain problems, for example, in imagery (Parmar et al. 2018) and audio (Child et al. 2019). Presser and Branwen (2020)

---

[1] rb.gy/dsdphc

[2] github.com/ElliotPenson/n-gram-chess

[3] github.com/thomasahle/fastchess

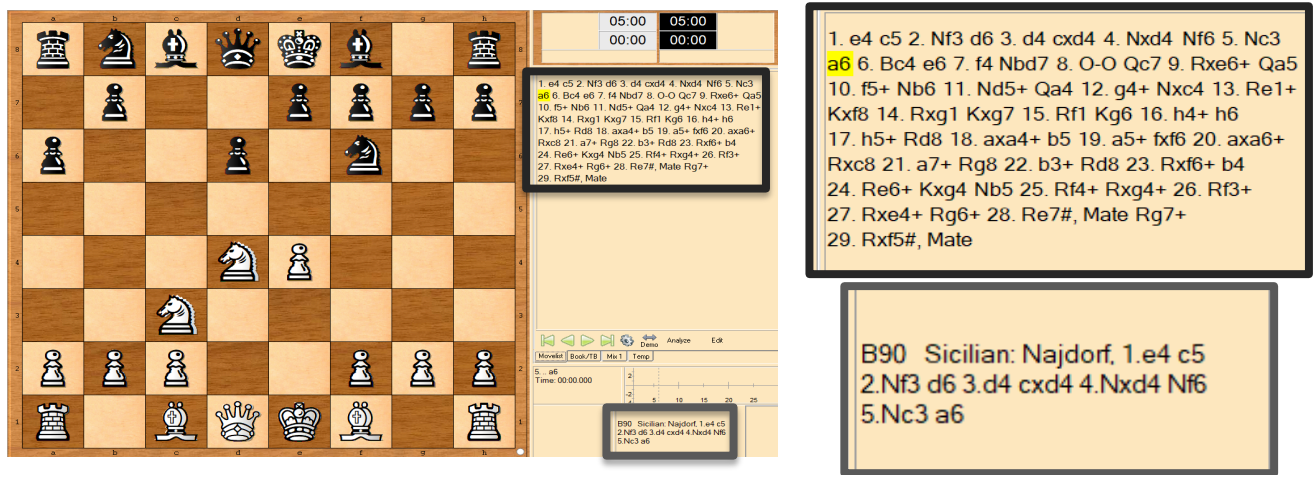

*Figure 1. Generated example games from GPT-2. Each transformer game generated is video captured and split for analysis move-by-move using Portable Game Notation (PGN) as inputs with both moving pieces and automated strategy notations*

first explored the application of GPT-2 to chess with a filter on invalid moves. Like recurrent neural networks, transformers however specialize in modeling sequential data like language (or game moves here), but without relying strictly on presentation order during training. Its core architecture features encoder-decoder cycles that apply weights to derive features with its unique 'attention' mechanism that effectively overweighs the most relevant features as it learns. The transformers' remarkable abilities to generate text arises from its parallelism during training, which enables traditional neural architectures to ingest vast amounts of internet-scale textual inputs. The chess application of GPT-2 suggests new and innovative ways to expand chess training data with high-level simulation data. For instance, our exploration of only high-ranking gameplay (e.g. training Elo ranks > 2,200) highlights the specialized transformer model may encode the basic features of multiple games and learn their winning strategies without human guidance or direction.

## Methods

### Datasets and Pre-processing

To generate chess training data for the language models, we transformed chess game archives to single lines beginning with either the simple result (win/lose/draw) or the result plus Elo ranks for black and white. In PGN format, the smaller dataset (milibrary.org)[4] consists of 11,291 games and over 800,000 player moves. For a game lasting an hour, the simulation learns from observing the equivalent to 10,000 hours of chess, roughly a year of continuous human play. The average Elo player rank of 1,815 was equally distributed between white and black with the average game length as 73 moves.

We also used the KingBase[5] dataset of 2.19 million PGN games archived as no older than 1990 and by players with Elo ranks greater than 2,000. Since almost 5,000 humans have an Elo ranking over 2,200 in 2016, this database presents language models with learning the move capabilities (without heuristics, hints, or rules) at the level of expert or candidate master.

The PGN move notation offers a loosely structured text file with metadata headers in square brackets followed by alternating black and white gameplay. The format offers a plain text training set that proves easy for humans to read or write, provides input to natural language modeling (transformers), and generates parsed games for machine-reading and display. The only header information needed for game replay proved to be the Result tag, so all other dates, player names, and locations were stripped for training purposes.

### Game Notation

To evaluate the success of the natural language model, just counting interesting PGN tokens gives a simple representation of what the model learns. The Arena game (Figure 1) interface illustrates the role of text in describing chess specifically. The algebraic move text notation (PGN) represents an equivalent token or word size ranging between 2-5 characters. The capture notation ("x"), numerical order of play (1…N), and piece abbreviation (K=king; Q=queen; R=rook, B=bishop; N=knight; P (or empty) = pawn). For knights, bishops, and rooks, their original location offers a unique

---

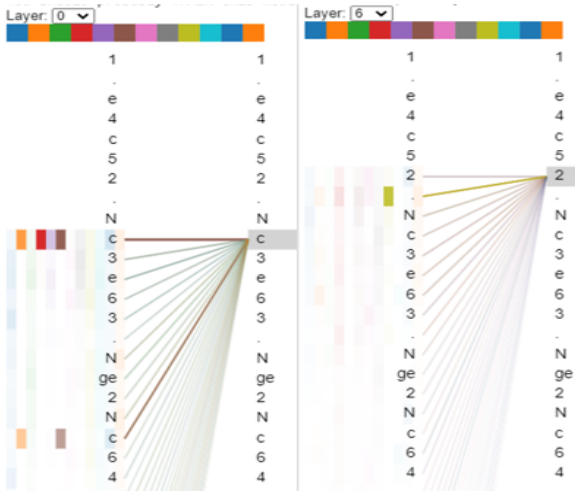

*Figure 2. Visualization of GPT-2 Attention Heads. Examples shown for layer 0 & 6 as samples.*

| Move | Frequency |
|------|-----------|
| Kingside Castle | 17.74% |
| Pawn Promotion | 13.67% |
| Check | 2.63% |
| Queenside Castle | 0.97% |

*Figure 3. Common Move Frequency Statistics of diverse example moves from 1000 games from the generative model.*

The evolution of transformers as sequence predictors follow on from problems encountered when training previous language models: 1) recurrent neural networks suffer from vanishing gradients; 2) long short-term memory (LSTM) networks predict unidirectionally with limited context and difficult parallelism. The transformer model introduced so-called "attention" which over-weights key parts of an entire sequence, train in parallel and process huge datasets without supervision or language labeling. As illustrated using GPT-2 visualization of chess moves in Figure 2, each token of the chess game receives varying importance through the transformers encoding-decoding layers (Vig 2019). For the present purposes, this core or universal language structure provides the pre-trained architecture and weights to fine-tune for many specialized tasks including chess game generation.

## Train Test Cycles

We perform 30,000 training steps and record plausible gameplay from unconditional (random) sampling without a prefix or trigger prompt. The leading token of a viable game to visualize begins with "[Result …] followed by whether the game gave a win to black ("0-1") or white ("1-0") or ended in a draw ("1/2-1/2"). Previous experiments using the small (124M) and medium (355M) models for GPT-2 for the game of Go motivated a starting point with the larger (774M) model. The extra-large (1.5 B) hyperparameter model was not tested owing to its large VRAM needs (>6 GB base model). These experiments follow the same progression of the small (124M) model being 4 times faster to train than the large one (774M). We ran multiple passes through the mixed, black-win, white-win, and draw datasets. Each run generated approximately 1000 games, with a 2-8% failure rate for non-viable gameplay or internally for illegal moves. We employed NVIDIA V100 GPUs (32 GB VRAM) as single units on a DGX supercomputer placement (Nge2 specifies Knight on the upper right g column, moving to e2). The destination of any move follows the standard game board locations (from white's perspective) with lower left (a1) to upper right (h8). Other specialized moves get custom notation: castling kingside (O-O) or queenside (O-O-O) and pawn promotions (appending an "=" and promotion piece to a queen from e8 is written e8=Q). Moves that put opponents in check ("+") or checkmate ("#") generate a sequence if pawn promotes to a queen (Q) for checkmate as e8=Q#, which also represent the 5-character (longest) token for the natural language model (other than "Result"). A key outcome of a successful language model would preserve the limited vocabulary and mimic the alternating player sequences making legal moves and generating strategic positions.

## Language Models

OpenAI's GPT-2 provides a convenient language model for text generation. We specialize in our approach using the Woolf version of the python package, gpt-2-simple[6]. In brief, GPT-2 represents a language model trained on 40 GB of text data as a corpus of highly ranked Reddit posts (Radford et al. 2018). The unidirectional training objective centers on predicting the next word (or token in a sequence), given all the previous ones. The second version from the original GPT scaled up the parameters (10x) and training data (10x), then showed promising signs of plausible machine text generation. One notable part of its 2-byte encodings also includes spacing, so the overall format of a given text document often is reproducible for poetry, film scripts, or other game notations.

---

[6] github.com/minimaxir/gpt-2-simple

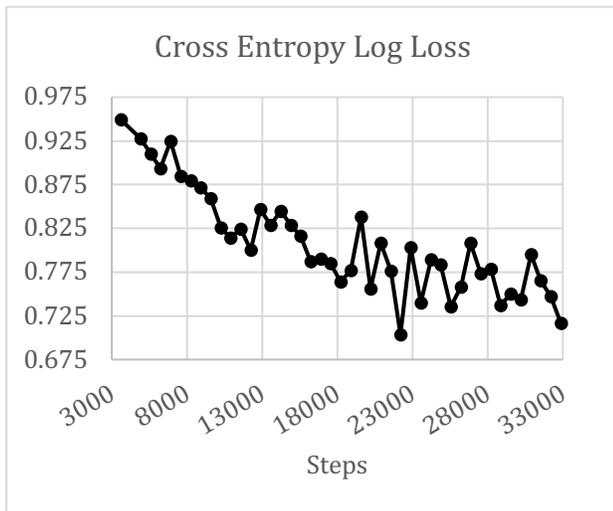

$$H_p(q) = -\frac{1}{N}\sum_{i=1}^{N} y_i \cdot log(p(y_i)) + (1 - y_i) \cdot log(1 - p(y_i))$$

Binary Cross-Entropy / Log Loss

*Figure 4. Training large GPT-2 with 2.8 million chess games in PGN text. The fine-tuned model uses the 774 million parameter (large) model with 33,000 training steps and achieves an average cross-entropy log loss value (smoothed = 0.72, final = 0.79).*

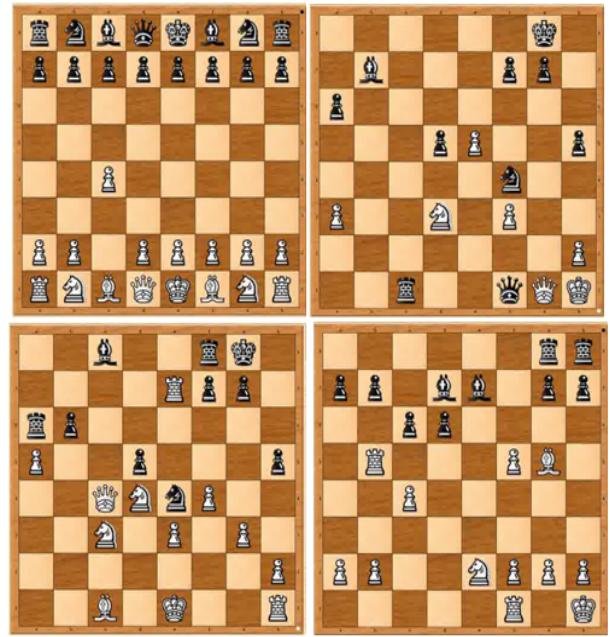

*Figure 5. Strategic positions from large GPT-2 generation. Upper left shows a classic English opening, upper right, a Slav Exchange; lower left, a King's Indian Defense (KID) and lower right, a Nimzowitsch Variation.*

(approximately 1 petaflop for 4x GPUs). Each run took approximately 8-10 hours for non-distributed training.

**Train Game Analysis Suite**

To analyze gameplay, we rely on the open-source Athena visualization for initial identifying interesting or classic game moves, particularly for commentary on opening moves (like "English Opening" or "Slav Exchange"). With a token counter (Figure 3), we analyze the frequency of some key moves learned by the language model: castling, pawn promotion ("="), check ("+"), and checkmate ("#").

**Results**

Even on the smaller subset (11k milibrary.org games) of expert training inputs, the large GPT-2 learns the input text with a cross-entropy log loss below 0.1, which generally represents high recall. This result compares to the more diverse but higher loss (0.72) for the same number of training steps (30k) but with the 10,000-fold larger Kingbase dataset (Figure 4). One extension of the current work is to evaluate the chess transformer by comprehensively compiling token frequencies for these classic strategic formations (e.g. simple counts of check moves, "+", or "O-O"). Because of the PGN notation, this statistical grading follows simply as token counting. To compare player success, we split apart all the wins into three classes of black wins, white wins, and draws. In this way we can train subsets of winning black players, then play those to get mock Elo rankings for the learning cases vs. random or mixed player colors. If an initial player enters the Chess Transformer at Elo base (800), losing 894 games successively against an expert with Elo 2,000 or greater would likely drop the white player to the floor rating of 100.

We also present this method to understand: 1) the transformer is learning a winning (and losing) side; 2) the effect of training data on outcomes. Generally, we find that after training against the smaller (milibrary.org) or larger (Kingbase) archives, both the chess transformers generate plausible gameplay. The PGN output particularly from the large archive and GPT-2 model mimics the Elo rank of 2,637, and equally balanced between white and black for 971 new games generated as valid. One interesting way to test if the chess transformer captures the syntax and meaning of PGN gameplay is to check for internal game consistency. For instance, do the generated games correctly show the higher-ranking Elo player as more likely to win? In 70.9% of generated games, the higher-ranked player (either white or black) is shown as the one who won. Some other simple tests include:

- Does it generate the back-and-forth play? (yes)
- Does it generate illegal moves? (yes, 10%)
- Does it generate impossibly repetitive play? (no)
- Does it mix up the ordinal play sequence? (no)

The average game length in the milibrary.org data is 73 moves, while for the generated games, we find a similar 67 moves per game. A key test of the chess transformer is whether it exhibits repetitive play (overfitting). Nearly one in five generated games shows at least one player castling to the kingside, with one in eight performing a pawn promotion.

In the Discussion section, we revisit the more strategic formations for the chess openings to understand if a coherent game approach follows from the transformer training. As an example, does the chess transformer generate more diverse openings than just the classic English Opening (1. c4)? With the game commentaries built into the Arena chess visualization, each replayed game from the chess transformer conveniently annotates strategic positions, while automatically moving the pieces according to the PGN we generate from the fine-tuned GPT language model. Notable examples shown in Figure 5 for these strategic positions do highlight with Arena chess: English opening (1. c4), Slav Exchange (4. Nc3 Nf6), King's Indian Defense (KID, 2. g3 Nf6 3. Bg2 Bc5) and a Nimzowitsch variation (2. Nc6). Not shown, are other noteworthy strategic movements generated from the chess transformer, such as Sicilian Defense (1. e4 c5); Reli Opening; Russian Game (Petroff Defense); Sicilian: Najdorf; English King's (2. Nc3 Nf8); English Four Knights (4. g3 d5 5. Cxd5); Neo-Old Indian (2. C4 e5 3. Nf3), King's Pawn Game; and Queen's Pawn Game.

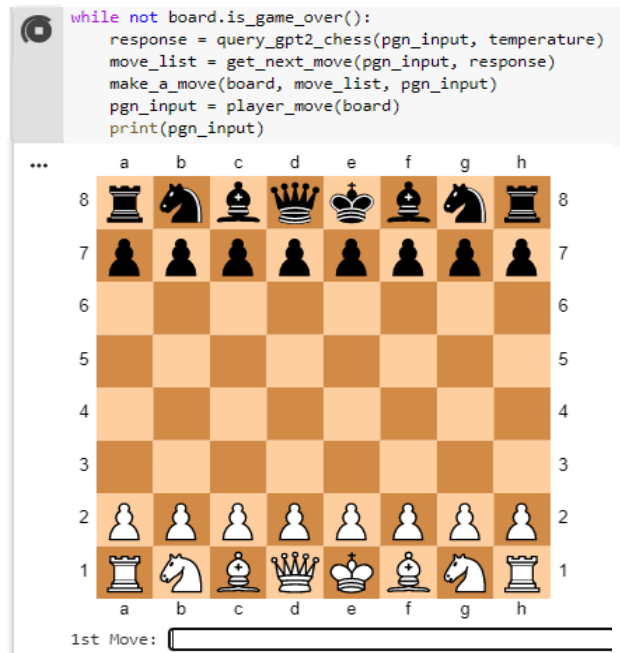

*Figure 2. Human vs. Machine in Live Play with Chess Transformer. The Colaboratory notebook includes pre-trained medium GPT-2 models and instructions for human (white) vs. machine (black) in head-to-head game play: rb.gy/dsdphc*

## Discussion

In addition to the current training methods, promising future work could segregate the training towards a biased subset and thereby contrast those specialized behaviors against the balanced white-black play described here. A similar course exists for comparing weak and strong players based on the subset of high and low ELO rankings. Since ELO is based on strength of opponent, one goal of that research centers on getting a rating for GPT-2 modeling itself. For example, does the smaller (124M) hyperparameter model underperform against the larger (774M) fine-tuned ones? While the current work focuses on plausible or interesting chess moves to learn in such a roundabout way, various other research paths could isolate the prompt or trigger phrases within GPT-2 finetuning and simulate better game rhythm from some midpoint in the game. As it stands, these simulations assume essentially a blank slate initial condition with no priors. The main result demonstrates that language modeling alone can generate plausible game playing by just observing the game and inferring all the rules and heuristics without manual interventions.

Two shortcomings of this work deserve attention. First, how is a language model able to learn chess? The GPT-2 has memorized a series of tokens and like any number of other statistical methods has figured out how to maximize rewards. Couldn't a Markov chain also string together enough tokenized abstract moves to convince the casual observer that the machine somehow comprehended the game? This big question segues to the second objection, which has featured prominently in the more advanced GPT-3 commentary. A common critique of transformers for text generation has centered on the simple hack test of "2+2=" or "2 + two ="? questions. In most cases, all transformers either repeat some joke from its training data (Reddit in the case of GPT) or otherwise mangled what seems obvious to even a second grader. In other words, are transformers mere parrots, or have they read sufficiently deep and far to capture the essence of a universal language model? Given the other multi-tasks applied to transformers, some challenges seem more compatible with the universal label. For example, topic summaries, question/answer, and one-shot translators provide some convincing and potentially profound insight into human and machine language. We investigate games in this realm of language archives for training data as an interesting example that also mirrors some of the strategic and interpretable aspects of language and knowledge understanding. Whether a human chess grandmaster has remarkable memory skills for previous winning moves seems largely irrelevant to their Elo rank or their formidable tournament play. Similarly, whether a 774 million hyperparameter language (or token) model has overfitted the game space for complex games like chess or Go, may also appear largely a philosophical question to ponder. The notion that a trained Elo 2,000 model consistently beats a lesser opponent suggests that perhaps there is a future better GPT-2 or GPT-3 text generator that can play chess at a consistent super-human level. It is worth noting finally that chess as a human endeavor (like Go) has largely been conceded to machine learning. Like calculating the square root of 23 or any other arithmetic operation, there is no real contest between how the human brain evolved and how a specialized calculator can perform in practice. What remains remarkable about creative gameplay however is the cross-over between those essential human traits, like language itself or creative inspiration, and the current progress in building massive transformer models.

A major challenge to the current approach is whether the GPT-2 gameplay is any good. Does the trained model advance beyond a parroted amateur? To examine this question systematically, the live gameplay interface (Figure 6) allows humans to play against the chess transformer and rate its overall effectiveness for themselves in match play. The live play relies on filtering approximately 10% of illegal moves using a Stockfish-inspired chess library (python-chess)[7]. In this paper, we have offered a few experimental criteria to answer other skill-related questions, such as segmenting training data into advanced or beginner examples to learn or looking for classic opening moves.

## Conclusion

Playing text-based games (including chess) proves possible by fine-tuning a custom GPT-2 language model. Using the PGN notation, full games of chess and its complex moves are cataloged in text. This represents another domain in which language models can be benchmarked against. Playing against the transformer reveals strong early gameplay as the number of strategies learned is large and weaker gameplay as the number of learned strategies falls significantly. While traditional game agents are trained with inherent game logic and MCTS's depth search, this approach highlights the notion that a breath search of millions of games can allow a language model to define a game's rules and strategy by itself.

## Acknowledgements

The authors would like to thank the PeopleTec Technical Fellows program for encouragement and project assistance. We are grateful to the front-line emergency workers who do their difficult work during the COVID-19 pandemic.

---

[7] github.com/niklasf/python-chess

## References

Bojanowski, Piotr, et al. "Enriching word vectors with subword information." Transactions of the Association for Computational Linguistics 5 (2017): 135-146.

Bory, P. (2019). Deep new: The shifting narratives of artificial intelligence from Deep Blue to AlphaGo. Convergence, 25(4), 627-642.

Ciolino, M., Noever, D. & Kalin, J. (2020). The Go Transformer: Natural Language Modeling for Game Play. arXiv preprint arXiv:2007.03500.

Child, R., Gray, S., Radford, A., & Sutskever, I. (2019). Generating long sequences with sparse transformers. arXiv preprint arXiv:1904.10509.

Devlin, J., Chang, M. W., Lee, K., & Toutanova, K. (2018). Bert: Pre-training of deep bidirectional transformers for language understanding. arXiv preprint arXiv:1810.04805.

Kirubarajan, A., & Dugan, L. (2020) Learning to Trick Humans: Evaluation Criteria for Human-Written and Computer-Generated Text. kirubarajan.nyc3.digitaloceanspaces.com/learning_to_trick_humans.pdf

Klein, T., & Nabi, M. (2019). Learning to Answer by Learning to Ask: Getting the Best of GPT-2 and BERT Worlds. arXiv preprint arXiv:1911.02365.

Lapan, M. (2018). Deep Reinforcement Learning Hands-On: Apply modern RL methods, with deep Q-networks, value iteration, policy gradients, TRPO, AlphaGo Zero and more. Packt Publishing Ltd.

Lee, J. S., & Hsiang, J. (2020). PatentTransformer-2: Controlling Patent Text Generation by Structural Metadata. arXiv preprint arXiv:2001.03708.

Müller, M., & Tegos, T. (2002). Experiments in computer Amazons. More Games of No Chance, 42, 243-260.

Nournai, C. F. (1997, July). Multiagent Chess Games. In Deep Blue Versus Kasparov: The Significance for Artificial Intelligence (pp. 45-52).

Parmar, N., Vaswani, A., Uszkoreit, J., Kaiser, Ł., Shazeer, N., Ku, A., & Tran, D. (2018). Image transformer. arXiv preprint arXiv:1802.05751.

Presser, S., Branwen, G. (2020) A Very Unlikely Chess Game, slatestarcodex.com/2020/01/06/a-very-unlikely-chess-game/

Qi, D., Su, L., Song, J., Cui, E., Bharti, T., & Sacheti, A. (2020). ImageBERT: Cross-modal pre-training with large-scale weak-supervised image-text data. arXiv preprint arXiv:2001.07966.

Radford, A., Wu, J., Child, R., Luan, D., Amodei, D., & Sutskever, I. (2019). Language models are unsupervised multitask learners. OpenAI Blog, 1(8), 9. github.com/openai/gpt-2

Silver, D., Schrittwieser, J., Simonyan, K., Antonoglou, I., Huang, A., Guez, A., ... & Chen, Y. (2017). Mastering the game of go without human knowledge. nature, 550(7676), 354-359.

Silver, D., Hubert, T., Schrittwieser, J., Antonoglou, I., Lai, M., Guez, A., ... & Lillicrap, T. (2017). Mastering Chess and Shogi by self-play with a general reinforcement learning algorithm. arXiv preprint arXiv:1712.01815.

Silver, D., Hubert, T., Schrittwieser, J., Antonoglou, I., Lai, M., Guez, A., ... & Lillicrap, T. (2018). A general reinforcement

learning algorithm that Masters Chess, Shogi, and Go through self-play. Science, 362(6419), 1140-1144.

Tang, R., Lu, Y., & Lin, J. (2019, November). Natural language generation for effective knowledge distillation. In Proceedings of the 2nd Workshop on Deep Learning Approaches for Low-Resource NLP (DeepLo 2019) (pp. 202-208).

Vaswani, A., Shazeer, N., Parmar, N., Uszkoreit, J., Jones, L., Gomez, A. N., ... & Polosukhin, I. (2017). Attention is all you need. In Advances in neural information processing systems (pp. 5998-6008).

Vig, J. (2019). OpenAI GPT-2: Understanding Language Generation through Visualization. Towards Data Science, via Medium, March, 5.

Vig, J. (2019). A multiscale visualization of attention in the transformer model. arXiv preprint arXiv:1906.05714. github.com/jessevig/bertviz